\documentclass[10pt,twocolumn,letterpaper]{article}

\usepackage[pagenumbers]{cvpr} 

\usepackage{algorithm}
\usepackage{algpseudocode}

\usepackage{amsmath}
\usepackage{amsfonts}
\usepackage{mathtools}
\newcommand{\R}{\mathbb{R}}
\newcommand{\nth}[1]{{#1_\text{th}}}

\newcommand{\C}{\mathcal{C}}
\newcommand{\Cs}{\mathcal{C}_\text{slide}}
\newcommand{\proc}{\mathcal{P}}
\newcommand{\imenc}{\mathcal{I}}

\newcommand{\Xm}[1]{X^{#1}}
\newcommand{\Xms}[1]{\tilde{X}^{#1}}
\newcommand{\Ym}[1]{Y^{#1}}
\newcommand{\Yms}[1]{\tilde{Y}^{#1}}
\newcommand{\Zms}[1]{\tilde{Z}^{#1}}

\newcommand{\imp}{\alpha}

\newcommand{\floor}[1]{\left\lfloor {#1} \right\rfloor}

\newcommand{\magnify}{\textsc{Magnify}}
\newcommand{\filter}{\textsc{Filter}}

\newcommand{\nmag}{n}
\newcommand{\npatch}{K}
\newcommand{\magf}{M}

\newcommand{\hackyparagraph}[1]{\vspace{3pt}\noindent\textbf{#1}}

\newcommand{\fullname}{\textbf{Pa}thology \textbf{T}ransformer with \textbf{H}ierarchical \textbf{S}election}
\newcommand{\name}{PATHS}

\usepackage{svg}

\usepackage{multirow}

\definecolor{cvprblue}{rgb}{0.21,0.49,0.74}
\usepackage[pagebackref,breaklinks,colorlinks,allcolors=cvprblue]{hyperref}

\def\paperID{11941} 
\def\confName{CVPR}
\def\confYear{2025}

\title{PATHS: A Hierarchical Transformer for Efficient Whole Slide Image Analysis}

\author{Zak Buzzard\textsuperscript{1}, Konstantin Hemker\textsuperscript{1}, Nikola Simidjievski\textsuperscript{2,1}, Mateja Jamnik\textsuperscript{1} \\
{\normalsize \textsuperscript{1}Department of Computer Science \& Technology, University of Cambridge, UK} \\
{\normalsize \textsuperscript{2}PBCI, Department of Oncology, University of Cambridge, UK} \\
{\tt\small [zzb20, kh701, ns779, mj201]@cam.ac.uk}
}

\begin{document}
\maketitle
\begin{abstract}
Computational analysis of whole slide images (WSIs) has seen significant research progress in recent years, with applications ranging across important diagnostic and prognostic tasks such as survival or cancer subtype prediction. Many state-of-the-art models process the entire slide -- which may be as large as $150,\!000\times150,\!000$ pixels -- as a bag of many patches, the size of which necessitates computationally cheap feature aggregation methods. However, a large proportion of these patches are uninformative, such as those containing only healthy or adipose tissue, adding significant noise and size to the bag.
We propose \fullname{} (\name), a novel top-down method for hierarchical weakly supervised representation learning on slide-level tasks in computational pathology. \name{} is inspired by the cross-magnification manner in which a human pathologist examines a slide, recursively filtering patches at each magnification level to a small subset relevant to the diagnosis. Our method overcomes the complications of processing the entire slide, enabling quadratic self-attention and providing a simple interpretable measure of region importance. We apply \name{} to five datasets of The Cancer Genome Atlas (TCGA), and achieve superior performance on slide-level prediction tasks when compared to previous methods, despite processing only a small proportion of the slide.

\end{abstract}    
\section{Introduction}

Whole slide images (WSIs) -- high resolution scans of sliced biopsy sections -- are the basis for pathologists to diagnose and analyse disease. Due to the importance and scale of this task, recent years have seen the development of a range of automated approaches to assist in processing and analysis, with particular success seen in the application of modern computer vision methods \cite{WSIDeepLearningSurvey}. However, the gigapixel scale of WSIs, coupled with their pyramidal structure, challenges the application of standard vision architectures such as convolutional neural networks \citep{ResNet, VGG16} and vision transformers~\citep{ViT} at the slide level. \looseness-1

When pathologists inspect whole slide images, they usually do so in a top-down manner: identifying regions of interest and tissue architecture (such as areas of cancerous tissue) at low magnification before investigating these areas further at greater magnification. To inspect the entire slide at its maximum resolution would be unduly time-consuming and largely \textit{uninformative}, with only certain areas of the slide providing useful information. Conversely, most state-of-the-art deep learning methods process the slide in its entirety at high magnification, splitting the image into a large collection of small (e.g., $256\times256$px) patches, in the order of magnitude of $10,\!000$ per slide \cite{WSIDeepLearningSurvey, MIL_survey_2018}. This incurs a high computational cost, and in many cases provides a large amount of uninformative data to the model, effectively creating a poor signal-to-noise ratio. Within this category, the most common approach is multiple instance learning (MIL), in which each slide is treated as a large unordered bag of patches that are processed using pre-trained computer vision models, and globally aggregated to produce slide-level representations \cite{MIL_survey_2018, MIL_ImageNet, ABMIL, MIL_pathology}. The global aggregation method must be efficient due to the scale of the bag; self-attention, for example, is infeasible, necessitating the use of less performant linear-time approximations \cite{TransMIL, nystrom}. Past approaches to mitigating computational overheads include selecting only a small proportion of patches by random selection~\cite{MIL_with_random_patch_selection} or 
manual clustering-based heuristic~\cite{MIL_strange_patch_subsampling}. However, such manual heuristics are suboptimal as they are error-prone and often inflexible. More recent work adapts hierarchical methods, which have seen success in the domain of computer vision, to WSIs \cite{nested_hierarchical_transformer, hierarchical_perceiver, Swin, wsi_hierarchical_vit}.
While more expressive than MIL, hierarchical methods nevertheless necessitate the pre-processing of the entire slide at its full magnification and require the use of self-supervision rather than task-specific training due to the large number of patches.  \looseness-1

\begin{figure*}[t]
    \centering
    \includegraphics[width=\linewidth]{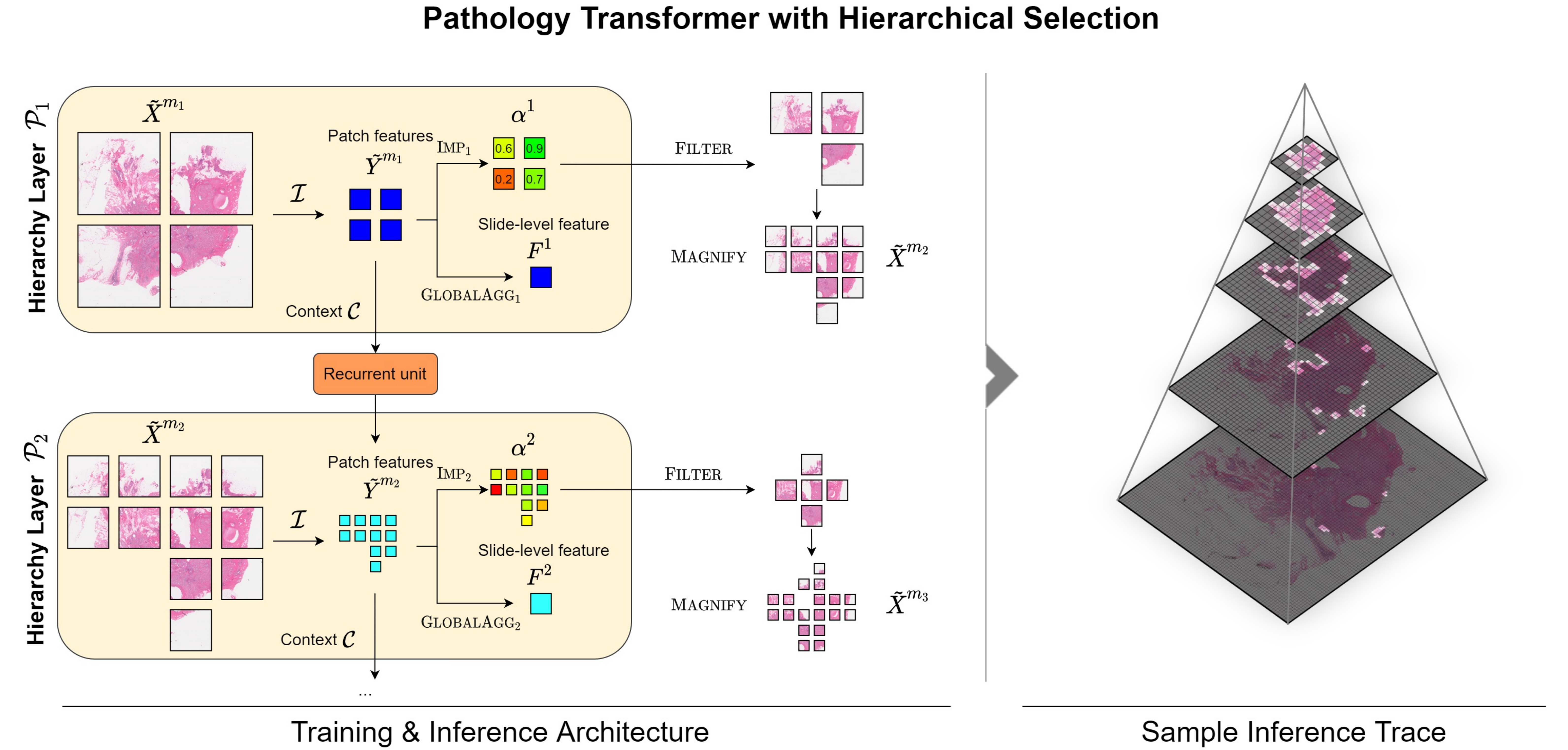}
    \caption{Overview of our novel method, \name{}, which predicts a patient's relative hazard level given a whole slide image using a \emph{top-down} hierarchical process along the slide's pyramidal structure, mimicking the workflow of a pathologist. The prediction $\hat{y}$ is made as a function of the slide-level features at each hierarchy level, $F^1, \dots, F^\nmag$.}
    \label{fig:intro_overview}
\end{figure*}
In this paper, we propose the \fullname{} (\name) -- a \emph{top-down} hierarchical model -- as a novel weakly supervised approach to learning on WSIs, combining the effectiveness of hierarchical methods with the data efficiency of patch sampling (summarised in \Cref{fig:intro_overview}). 
Much like a pathologist, our model initially processes the slide at a low magnification, capturing high-level tissue structure, before an attention mechanism recursively identifies regions of importance. The regions of highest importance are magnified and the process is repeated, retaining information from lower magnifications in the form of a hierarchy. This enables the capture of information across a range of resolutions while avoiding the costly processing of the entire slide. We show that \name{} exhibits several desirable properties for slide-level tasks in computational pathology: 

\begin{itemize}
    \item \textbf{High accuracy} on five WSI datasets, covering different cancer sites, achieving comparable or improved performance on a survival prediction task compared to state-of-the-art methods. Our proposed dynamic patch selection and multi-resolution slide context drive this performance, reading in fewer uninformative slides at each magnification and thus improving the signal-to-noise ratio.
    \item \textbf{Computational efficiency} by only processing a fraction of the slide at each magnification, leading to a speed-up exceeding a factor of ten in inference time at $10\times$ magnification compared to MIL. 
    \item A clinically \textbf{meaningful heuristic} for patch selection that mimics the workflow of a pathologist.
    \item Top-down patch selection can be used for \textbf{debugging and validation}. Explicit identification of regions of interest enables visualisation of the learned traversal through the WSI's hierarchical structure, and provides interpretable model behaviour.
\end{itemize}

\section{Related Work}
\hackyparagraph{Multiple Instance Learning} Whole slide images store scans of a slide at several magnifications, the highest of which corresponds to an image of up to $150,\!000\times150,\!000$px. Due to this large scale, multiple instance learning (MIL) \cite{MIL_survey_2018, MIL_pathology, MIL_with_random_patch_selection} is frequently used in computational pathology tasks. Multiple instance learning treats each slide as a large unordered bag of low-resolution patches (e.g., $256\times256$ px) at a fixed magnification level.

General-purpose MIL approaches include ABMIL~\cite{ABMIL}, which introduces an attention-based aggregation as a global weighted sum, where the per-patch weights are scalars produced as a learnable function of the patch feature.  
Given the success of self-attention in the domain of vision \cite{ViT, Swin}, several works have explored self-attention based MIL aggregation~\cite{TransMIL}.
However, in the context of computational pathology, the scale of the WSIs precludes the use of full self-attention, due to quadratic scaling with respect to the sequence length \cite{Transformer}, forcing these methods to use less performant compromises such as linear-time approximations~\cite{nystrom} or cross-attention with a smaller set~\cite{multimodal_coattn_transformer}.

To mitigate the issues caused by the large scale of WSIs, related work has focused on reducing the bag size through random or heuristic-based sampling, or the clustering of patches into smaller bags~\cite{MIL_with_random_patch_selection, MIL_strange_patch_subsampling, DeepAttnMIL}. Graph neural networks have seen use as an aggregator of randomly sampled patches, accounting for spatial interactions~\cite{Li2018GraphCF, Zhao2020PredictingLN}. However, these non-parametric patch sampling methods risk missing important sections of the slide, and may fail to adequately represent large-scale features. More recently, \citet{thandiackal2022zoommil} propose ZoomMIL, a cross-magnification MIL method which selects patches in a learnable manner through  an adapted differentiable patch selection algorithm~\cite{cordonnier2021differentiable}, removing the need for manual heuristics. The benefits of incorporating patch features from multiple magnification levels has been observed in other work, demonstrating the potential of this technique to improve slide-level representations in MIL \cite{multi_res_MIL, Li2020DualstreamMI}. Regardless, these methods remain limited by the set-based nature of MIL, in which the overall structure of the slide, and spatial relationships between patches, are lost due to the discarding of positional information.

\hackyparagraph{Hierarchical Methods} Hierarchy-based image processing enables positional contextualisation of image patches and processing across multiple image scales, extending efficiently to large images \cite{hierarchical_perceiver, nested_hierarchical_transformer, Swin}. Rather than globally aggregating the image in one step, as in most MIL approaches, the grid of patches is repeatedly aggregated across spatially local steps. The resulting features form a hierarchy, where higher levels represent larger regions of the image, with the topmost level containing a single global feature.

In the context of computational pathology, \citet{wsi_hierarchical_vit} propose Hierarchical Image Pyramid Transformer (HIPT), which improves expressivity over the standard set-based MIL methods, enabling the capture of macro-scale features and large cell structures in the slide. However, as training on entire slides in an end-to-end manner is computationally infeasible, training is split into several distinct stages, each corresponding to a single magnification and level of the hierarchy. Due to a lack of patch-level labels, all but the last stage must be pre-trained using a self-supervised method \cite{DINO}, which could lead to the inclusion of redundant visual information (such as scanning artefacts, background proportion, biopsy shape). More notably, however, these approaches operate in a \textit{bottom-up} manner: the slide is initially processed as a grid of patches at the highest magnification (the bottom level of the hierarchy), with subsequent processing moving up the hierarchy to the lower magnification levels. This necessitates costly processing of a large number of patches per slide, likely including many of low relevance to the downstream task. We, therefore, propose a \emph{top-down} approach, which retains the hierarchical structure, while iteratively selecting substantially smaller but important areas of the slide.

\section{Method}

\subsection{Notation}
Given a WSI $X$, let $\Xm{m}$ denote the collection of (square, non-overlapping) patches of size $s\times s$ at magnification $m$, indexed by position $(u,v)$, so $\Xm{m}_{u,v} \in \R^{s\times s\times 3}$. Patches are processed by a pre-trained image encoder $\imenc$, such that $\imenc(\Xm{m}_{u,v})\in\R^{d}$ for some dimension $d$. We consider an arbitrary weakly-supervised task, with the goal of modelling a distribution $p(y\mid X)$ (e.g., survival prediction).

\subsection{Patch Selection}
At each magnification level $m$ we identify a small subset of patches $\Xms{m} \subseteq \Xm{m}$ to process. Unlike previous methods, which define $\Xms{m}$ non-parametrically using random choice or manual heuristics \cite{MIL_with_random_patch_selection, MIL_strange_patch_subsampling}, \name{} enables such a subset to be selected by the model during training. We achieve this by processing patches at $\nmag$ magnification levels $m_1 < m_2 < \dots < m_\nmag$, which form a geometric sequence, $m_{i+1} = \magf m_i$, to ensure patch alignment between levels. The model consists of $\nmag$ processors $\proc_1, \proc_2, \dots, \proc_\nmag$, the $\nth{i}$ of which is dedicated to processing patches at magnification $m_i$. $\proc_i$ additionally learns a scalar importance value $\alpha^i_{u,v} \in [0, 1]$ for each patch $\Xms{m_i}_{u,v}$, which models the relative importance of the patch, and provides a learnable heuristic for patch selection at the subsequent level:
\[
\Xms{m_1}=\Xm{m_1}
\]
\begin{equation}
\Xms{m_{i+1}} = \magnify(\filter(\Xms{m_i}, \alpha^i)).
\end{equation}
\filter{} retains only the $\npatch$ patches of highest importance, where $\npatch$ is a hyperparameter. \magnify{} queries the WSI in the same location as these patches, but at the subsequent resolution, effectively `zooming in' on the selected patches, then removing resultant patches which consist only of background. This process is visualised in \Cref{fig:intro_overview}.

As patch size (in pixels) is kept constant across each hierarchy level, magnification produces $\magf^2$ output patches for each input (or fewer when background is present). As a result, we have a fixed upper bound of
\begin{equation} \label{eq:patch_bound}
|\Xms{m_i}| \leq \magf^2 \npatch
\end{equation}
for $i>1$. We use $\magf=2$ in all experiments to enable a larger value of $\npatch$. By choosing a low starting magnification $m_1$, we also ensure that $\Xms{m_1}=\Xm{m_1}$ contains a small number of patches.

As the predicted values of $\alpha^i$ change during training, this technique effectively exposes the model to a large number of distinct patches over the course of training (regardless of $\npatch$), helping to avoid overfitting.

\subsection{Context}
At higher magnification levels, only a small fraction of the slide's total area is visible to the model, making it beneficial to pass on information from prior magnification levels. We refer to this information as \textit{context}, allowing the model to account for macro-scale slide properties when processing patches at high magnification, and employ it at both the patch- and slide-level.

\hackyparagraph{Hierarchical Context} Patch-level hierarchical context informs the model of the nature of the tissue surrounding each patch. This allows the incorporation of high-level features, such as tumour size, into the representations of patches at high magnification.

For each patch $\Xm{m_i}_{u, v}$ at magnification $m_i$, at each prior magnification level $m_j$ ($j<i$) there is a unique `parent' patch at position $(u_j, v_j)$ such that the slide area covered by patch $\Xm{m_j}_{u_j,v_j}$ includes that of $\Xm{m_i}_{u, v}$. We define the hierarchical context of a patch $\Xm{m_i}_{u,v}$ as the list of all patch embeddings from parent patches at previous magnification levels,
\begin{equation}
\C\left(\Xm{m_i}_{u,v}\right) = [\imenc(\Xm{m_1}_{u_1,v_1}), \dots \imenc(\Xm{m_{i-1}}_{u_{i-1},v_{i-1}})].
\end{equation}

$\C$ provides context for an individual patch by representing the surrounding area of the slide.

\hackyparagraph{Slide-level Context} In addition to hierarchical patch-level context, we find it beneficial to pass high-level global information between magnification levels. To achieve this, each processor $\proc_i$ produces a slide-level (but magnification specific) representation $F^i$ following global aggregation.

Then, rather than considering the final feature $F^{\nmag}$ only, the final target prediction is modelled as a function of the slide context $p(y\mid \Cs)$, where
\begin{equation}
\Cs = [F^1, \dots, F^\nmag].
\end{equation}

In our experiments we carry out a simple summation reduction over the slide-level context, $F=\sum_i F^i$, followed by a single linear layer to produce $\hat{y}$, leading to a residual model in which each processor after the first models an offset for the global feature. We leave exploration of more complex aggregation of the cross-magnification features $\Cs$ to future work.

\subsection{Processor Architecture}
Each processor $\proc_i$ consists of a contextualisation module, which incorporates hierarchical context into patch features, a transformer-based global aggregator, and an importance modelling module.
Conditioned on the patches $\Xms{m_i}$, and per-patch hierarchical context $\C(\Xms{m_i})$, each processor produces an aggregated feature and importance predictions,
\begin{equation}
F^i, \alpha^i = \proc_i(\Xms{m_i}, \C(\Xms{m_i})).
\end{equation}


\hackyparagraph{Contextualisation Module}
\begin{figure}[t]
    \centering
    \includegraphics[width=\linewidth]{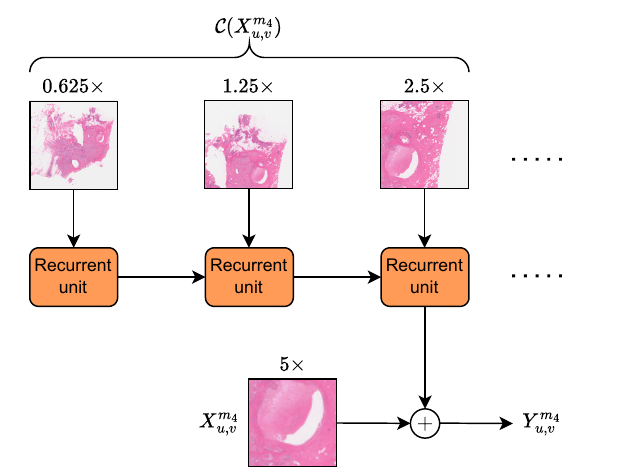}
    \caption{Architecture of the contextualisation module, which accounts for the hierarchical context of a patch $\Xm{m}_{u,v}$. The recurrent units are applied down the hierarchy, forming a tree-shaped RNN. In this example, $m_1=0.625$ and $\magf=2$.}
    \label{fig:contextualisation_module}
\end{figure}
\Cref{fig:contextualisation_module} illustrates the architecture of the contextualisation module. At high magnification, each patch feature contains information localised to an extremely small section of the slide; contextualisation aims to adapt these features to incorporate macro-scale tissue information.
For a patch $\Xm{m_i}_{u,v}$, the contextualised feature $\Ym{m_i}_{u,v}$ is defined as
\begin{equation}
\Ym{m_i}_{u,v} = \imenc(\Xm{m_i}_{u,v}) + \text{RNN}(\C(\Xm{m_i}_{u,v})),
\end{equation}
where RNN denotes a learnable recurrent neural network, which is applied sequentially to the hierarchical context list $\C(\Xm{m_i}_{u,v})$. In this manner the RNN produces a feature offset which accounts for high-level properties of the tissue surrounding each patch, thus `contextualising' the patch feature. Summation of the RNN output was chosen to enable easy representation of the identity function $\Ym{m_i}_{u,v}=\imenc(\Xm{m_i}_{u,v})$, for cases in which a patch's surrounding tissue is not of high relevance.

By sharing the weights of the RNN between all processors, this operation may be implemented efficiently: each processor carries out a single recurrent unit update step per patch, passing the resulting state to the corresponding patches at the subsequent magnification level.

\hackyparagraph{Importance Modelling} To enable patch selection, each processor $\proc_i$ implicitly learns scalar importance values $\imp^i$ for patches at magnification $m_i$. This is achieved through a gating mechanism, in which a two-layer MLP followed by sigmoid activation (denoted $\textsc{Imp}_i$) is applied to the contextualised patch embeddings $\Yms{m_i}$, producing scalar weights. Each embedding is then scaled by its corresponding weight to produce the final set of features $\Zms{m_i}$,
\begin{equation}
\Zms{m_i}_{u,v} = \alpha^i_{u,v} \Yms{m_i}_{u,v}.
\end{equation}
These features are globally aggregated, causing the model to assign higher importance values to patches with greater information content, as observed in past work \cite{ABMIL, MultimodalDynamics, multi_res_MIL}.

\hackyparagraph{Global Aggregation} Following the success of self-attention based aggregation \cite{TransMIL, wsi_hierarchical_vit, multimodal_coattn_transformer}, the contextualised, importance scaled patch features $\Zms{m_i}$ are aggregated globally via a transformer decoder (denoted $\textsc{GlobalAgg}_i$). We incorporate a two dimensional positional encoding (based on that of \citet{Transformer}) due to the sparse distribution of patches across the slide's area.
Aggregation produces the slide-level feature $F^i$ for magnification level $i$, which is added to the slide-level context $\Cs$.


\Cref{alg:processor} summarises the procedure carried out by each processor $\proc_i$, and the overall method for processing a slide $X$ using \name{} is summarised in \Cref{alg:slide_proc} (for both, see \Cref{app:algorithms}).

\section{Experiments}

\hackyparagraph{Datasets} The Cancer Genome Atlas (TCGA) provides public databases of documented cancer cases over a range of sites, including diagnostic whole-slide images among other data. We evaluate \name{} on the survival prediction task across five cancer types: IDC (invasive ductal carcinoma), CRC (colorectal cancer), CCRCC (clear cell renal cell carcinoma), PRCC (papillary renal cell carcinoma) and LUAD (lung adenocarcinoma), which we select due to their large size within TCGA and frequent use in past work. We cross-validate our method across five folds for each dataset, using the same folds for each model.

\begin{table*}[t]
    \centering
    \advance\leftskip-3cm
    \advance\rightskip-3cm
       \caption[Main results table]{C-index performance on the survival prediction task cross-validated over the same five folds, including sample standard deviation across folds. \name{} achieves superior performance on four out of five datasets, and the highest overall performance.} 
    \begin{tabular}{l|ccccc|c}
         \toprule
         Architecture & IDC & CRC & CCRCC & PRCC & LUAD & Mean \\
         \midrule
         ABMIL \cite{ABMIL} & $0.487\pm0.079$ & $0.566\pm0.075$ & $0.561\pm0.074$ & $0.671\pm 0.076$ & $0.584 \pm 0.054$ & 0.574 \\
         DeepAttnMISL \cite{DeepAttnMIL} & $0.472\pm 0.023$ & $0.561\pm0.088$ & $0.521\pm0.084$ & $0.472 \pm 0.162$ & $0.563 \pm 0.037$ & 0.518 \\
         GCN-MIL \cite{Li2018GraphCF, Zhao2020PredictingLN} & $0.534 \pm 0.060$ & $0.538\pm0.049$ & $0.591\pm0.093$ & $0.636 \pm 0.066$ & $\bf 0.592 \pm 0.070$ & 0.578 \\
         DS-MIL \cite{Li2020DualstreamMI} & $0.472 \pm 0.020$ & $0.470\pm0.053$ & $0.548\pm0.057$ & $0.654 \pm 0.134$ & $0.537 \pm 0.061$ & 0.536 \\
         HIPT \cite{wsi_hierarchical_vit} & $0.634 \pm 0.050$ & $ 0.608\pm0.088$ & $ 0.642\pm0.028$ & $0.670 \pm 0.065$ & $0.538 \pm 0.044$ & 0.618 \\
         ZoomMIL \cite{thandiackal2022zoommil} & $0.588 \pm 0.062$ & $0.631 \pm 0.065$ & $0.647 \pm 0.069$ & $0.662 \pm 0.076$ & $0.551 \pm 0.051$ & 0.616 \\ \midrule
         \name{} (ours) & $\bf 0.636\pm0.069$ & $\bf 0.695\pm0.097$ & $\bf 0.677\pm0.046$ & $\bf 0.772\pm0.036$ & $0.545\pm0.060$ & \bf 0.665 \\
         \bottomrule
    \end{tabular}
 
    \label{tab:main_results}
\end{table*}

\hackyparagraph{Baselines}
We compare \name{} to a number of state-of-the-art weakly-supervised baselines:

\begin{itemize}
    \item \textbf{ABMIL} \cite{ABMIL}: Attention-Based Multiple Instance Learning (ABMIL) is a simple MIL variant. Scalar attention values are produced per patch, and used as weights in a global sum for slide-level aggregation.
    \item \textbf{DeepAttnMISL} \cite{DeepAttnMIL}: Variant of ABMIL, with the addition of phenotype-based clustering.
    \item \textbf{GCN-MIL} \cite{Li2018GraphCF, Zhao2020PredictingLN}: A GNN-based MIL approach. The slide is processed by several graph convolution layers before aggregation.
    \item \textbf{DS-MIL} \cite{Li2020DualstreamMI}: A MIL-based approach employing contrastive learning, multiple magnifications and a modified aggregation function.
    \item \textbf{HIPT} \cite{wsi_hierarchical_vit}: Hierarchical Image Pyramid Transformer (HIPT) aggregates the entire slide in three vision transformer-based hierarchical stages. The bottom two stages are trained in a self-supervised manner using DINO \cite{DINO}. Due to its hierarchical nature, we consider this baseline an important comparison for our work.
    \item \textbf{ZoomMIL} \cite{thandiackal2022zoommil}: A MIL approach in which patches are selected from multiple magnifications via a differentiable zooming procedure. We configure ZoomMIL to sample the same number of patches as \name{} at each magnification for fair comparison (further details in \Cref{app:hyperparameters}). \looseness -1
\end{itemize}
While all models are evaluated on the same folds and datasets, the results for \mbox{ABMIL}, DeepAttnMISL, GCN-MIL, DS-MIL and HIPT use pre-calculated risk scores for these folds, as reported in~\cite{wsi_hierarchical_vit}.

\hackyparagraph{Setup} It is common to process whole slide images at $10\times$ or $20\times$ magnification to capture the details of individual cells \cite{wsi_hierarchical_vit, ABMIL, multimodal_coattn_transformer, MIL_ImageNet}. In all experiments, we select $10\times$ magnification as the bottom level of the hierarchy $m_\nmag$, and $m_1=0.625\times$ as the top, ensuring that $\Xms{m_1} = \Xm{m_1}$ is of tractable size of all slides, leading to five magnification levels. We also set $\npatch=20$, causing a fixed limit of $\magf^2 \npatch = 80$ patches per slide at each magnification, a small fraction of the total (which may be as many as tens of thousands).

To train the model for survival prediction we use the censored negative log-likelihood training objective $\mathcal{L}_{NLL}$ \cite{NLL_Loss} with $\alpha=0.6$. We quantise patient survival times into $b$ buckets such that each bucket contains roughly an equal number of uncensored patients. The model outputs $b$ logits, corresponding to the survival hazards for each bucket, from which $\mathcal{L}_{NLL}$ may be computed. We set $b=4$ in all experiments.

We evaluate using the censored concordance index metric (c-index), which measures the proportion of comparable patient pairs (those in which one can tell with certainty the order in which the events occurred) for which the model's survival prediction is concordant, as is standard. Random choice achieves a score of $0.5$, while the best possible score is $1$. All experiments were run on a single Nvidia A100 80GB GPU. See \Cref{app:hyperparameters} for a complete list of hyperparameters. The code to reproduce the experiments is available at \url{https://github.com/zzbuzzard/PATHS}. \looseness -1

\hackyparagraph{Patch Embedding} We pre-process all patches using a pre-trained image encoder $\imenc$, avoiding the heavy I/O and computation cost of reading and processing the patches during training. The results are stored as a two dimensional array of features, rather than an unordered bag as in MIL, to preserve positional information. Furthermore, unlike traditional MIL techniques, we must pre-process patches at several magnification levels, rather than at the highest magnification only: the total number of patches to be pre-processed per slide is $\sum_{i=1}^\nmag |\Xm{m_i}|$ rather than $|\Xm{m_\nmag}|$. However, note that $|\Xm{m_n}|, |\Xm{m_{n-1}}|, \dots$ forms a geometric sequence, as each time magnification is reduced by a factor of $\magf$, the number of patches $|\Xm{m}|$ falls by a factor of $\magf^2$. In the case of $\magf=2$, which we use in all experiments, our method incurs a pre-processing overhead of a factor of $1+\frac14+\dots+\frac1{4^{n-1}}\leq\frac{4}{3}$. Note that this overhead is only required to accelerate training, and during inference only the selected patches $\Xms{m}$ are extracted from each level. This is in contrast to past work, in which a new slide must be fully patched and pre-processed before inference begins, incurring high latency.

In this work, we employ UNI \cite{chen2024uni} as our patch encoder~$\imenc$. UNI is a recent vision transformer, pre-trained on a large dataset of WSI patches, excluding datasets used in our evaluation (such as TCGA) to prevent data contamination. Comparison to alternative encoders can be found in \Cref{app:image_encoder_ablation}. \looseness -1

\section{Results}
Table \ref{tab:main_results} shows the performance of our model \name{} against several baselines on the survival prediction task. 
\name{} achieves the highest overall c-index across the five cancer subtypes, with the highest performance on four of the five datasets, despite processing only a small fraction of the slide. Compared to ABMIL, DeepAttnMISL, GCN-MIL, DS-MIL and HIPT, all of which process the entire slide as tens of thousands of patches at $20\times$ magnification, \name{} processes just several hundred patches per slide. Despite this, we achieve a significant improvement in model accuracy, highlighting the benefit of processing a smaller number of more relevant patches. The improvement over \mbox{ZoomMIL}, which similarly filters the patches to a small subset per slide, demonstrates the advantage of \name{} over MIL architectures.

\begin{figure}[b]
    \centering
    \includegraphics[width=0.9\linewidth]{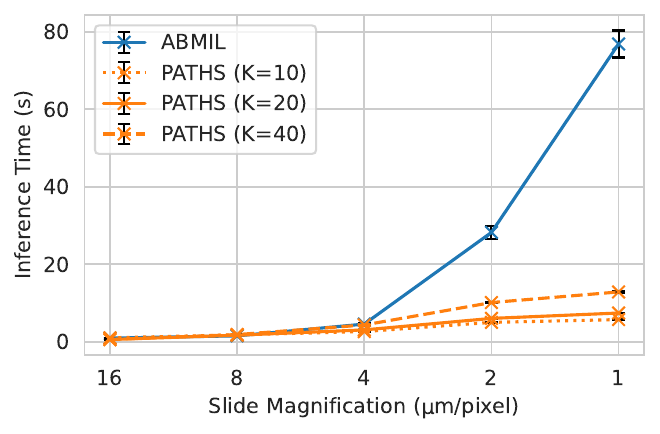}
    \caption{Inference speed, including I/O, patch pre-processing using UNI (which dominates latency), and model inference of \name{} (orange) compared to ABMIL (blue) when applied to a single new WSI. The magnification levels shown correspond to those in our experiments ($m_5=10\times=1\mu\text{m}/\text{pixel}$). As pre-processing dominates latency, the results for ABMIL are very close to those for other full slide baselines. Values were averaged over 50 TCGA-BRCA slides on a high performance A100 workstation, with standard error of the mean shown. The results clearly show the low latency of \name{} compared to methods which process the full slide, even for larger values of $\npatch$.}
    \label{fig:inference_speed}
\end{figure}

\hackyparagraph{Inference Speed} When vision models are incorporated into practical tools for computational pathology, it is imperative to achieve low computational overhead and inference latency, since  computational resources are often limited in a clinical setting. Whilst large-scale offline preprocessing of patch features enables fast training for `full slide' methods (i.e., those which must process all tissue-containing patches at high magnification, such as ABMIL or HIPT), this workaround does not extend to inference time. When applied to a new slide in a clinical setting, the entire slide must first be loaded into memory and processed using the patch embedding network (which may be a large network, such as UNI), leading to significant latency even on high performance infrastructure. \Cref{fig:inference_speed} demonstrates that, by significantly reducing the number of patches required from each slide, \name{} significantly improves inference latency over full slide approaches. This is a key advantage of our method, as this preprocessing step is the dominant processing cost for both \name{} and full slide models at inference time, taking up over 99\% of inference time in our experiments. Note that, even on state-of-the-art hardware and at just $10\times$ magnification (while $20\times$ is common), a minimal full slide approach takes over a minute to process a single new slide on average. It is reasonable to assume that latency will be significantly larger in practice, especially in the case of models running locally on clinical hardware to ensure patient confidentiality. \Cref{app:inference_speed_details} provides further details on the number of patches loaded by each approach (which is roughly proportional to inference latency) for a hardware-independent comparison of efficiency.

\begin{table*}
    \centering
    \advance\leftskip-1cm
    \advance\rightskip-1cm
        \caption{Ablation study: in order to demonstrate the efficacy of patch contextualisation, slide-level context and attentional patch selection, we compare our model against several simpler variants on the survival prediction task across all datasets. We show that each module contributes to the overall performance.}
        \begin{tabular}{c|ccc|c|c}
         \toprule
        & \multicolumn{3}{|c|}{Context mode} & \multirow{2}{*}{Random patch selection} & \multirow{2}{*}{\name} \\
        & Neither & Hierarchical only & Slide-level only & & \\
         \midrule
IDC & $0.607 \pm 0.047$ & $0.618 \pm 0.048$ & $\bf 0.641 \pm 0.058$ & $0.608 \pm 0.080$ & $0.636 \pm 0.069$\\
CRC & $0.644 \pm 0.058$ & $0.641 \pm 0.131$ & $0.687 \pm 0.085$ & $\bf 0.699 \pm 0.086$ & $0.695 \pm 0.097$\\
CCRCC & $0.642 \pm 0.056$ & $0.646 \pm 0.051$ & $0.670 \pm 0.042$ & $0.664 \pm 0.010$ & $\bf 0.677 \pm 0.046$\\
PRCC & $0.705 \pm 0.082$ & $0.769 \pm 0.028$ & $0.727 \pm 0.069$ & $0.768 \pm 0.047$ & $\bf 0.772 \pm 0.036$\\
LUAD & $\bf 0.573 \pm 0.070$ & $0.549 \pm 0.071$ & $0.557 \pm 0.049$ & $0.539 \pm 0.073$ & $0.545 \pm 0.060$\\
\midrule
Mean & 0.634 & 0.645 & 0.656 & 0.656 & \bf 0.665\\         \bottomrule
    \end{tabular}

    \label{tab:ablation_without_stuff}
\end{table*}

The main novelties of \name{} are the learnable patch selection module, combined with the patch- and slide-level context, allowing the propagation of cross-magnification information down the hierarchy. To investigate the contribution of each module to the overall performance of \name{}, we carry out an ablation study (\Cref{tab:ablation_without_stuff}) in which we evaluate several variants of our architecture on the five datasets used in \Cref{tab:main_results}. \looseness -1

\hackyparagraph{Cross-magnification Context Improves Over MIL} With both hierarchical and slide-level context removed, our model becomes similar to a single magnification MIL method. The drop in performance highlights the advantage of our method over MIL, although the score remains relatively strong across the datasets, likely due to the strength of transformer-based aggregation over a small set of extracted relevant patches. The addition of either hierarchical or slide-level context further improves performance, particularly that of slide-level context, demonstrating the benefit of incorporating cross-magnification information, as observed in other work \cite{multi_res_MIL,  wsi_hierarchical_vit, Li2020DualstreamMI}. However, it should be noted that for the LUAD dataset, on which \name{} performs poorly, the removal of context leads to \textit{improved} performance. As both ZoomMIL and HIPT also perform poorly on LUAD (\Cref{tab:main_results}), we hypothesise that cross-magnification information may be of low importance on this particular dataset and task, as evidenced by the strong performance of single magnification methods such as ABMIL. \label{sec:ablation}

\hackyparagraph{Benefit of the Learned Sampling Heuristic} Next, we investigate the significance of extracting patches based on the predicted importance $\alpha$. This is achieved through the replacement of the importance MLP ($\textsc{IMP}_i$) with a random distribution $\alpha^i_{u,v}\sim U[0,1]$ at inference time, leading to the selection of random areas of the slide (although background patches are still excluded). Interestingly, this modification leads to only a small reduction in performance. This result is supported by past work: \citet{MIL_with_random_patch_selection} achieve reasonable performance in a multiple instance learning pipeline using just 16 random patches from each slide, which the authors argue are very likely to contain at least one relevant (e.g., tumorous) patch. Our method uses both a higher number of patches (at most 80 per level) and multiple magnification levels, greatly increasing the likelihood of capturing relevant information under random selection. However, random sampling foregoes the interpretability benefits of attentional sampling, which allow us to easily inspect model behaviour.

\begin{figure}[t!]
\centering
\begin{subfigure}[t]{0.5\textwidth}
    \includegraphics[trim={0 25 0 25},clip,width=\textwidth]{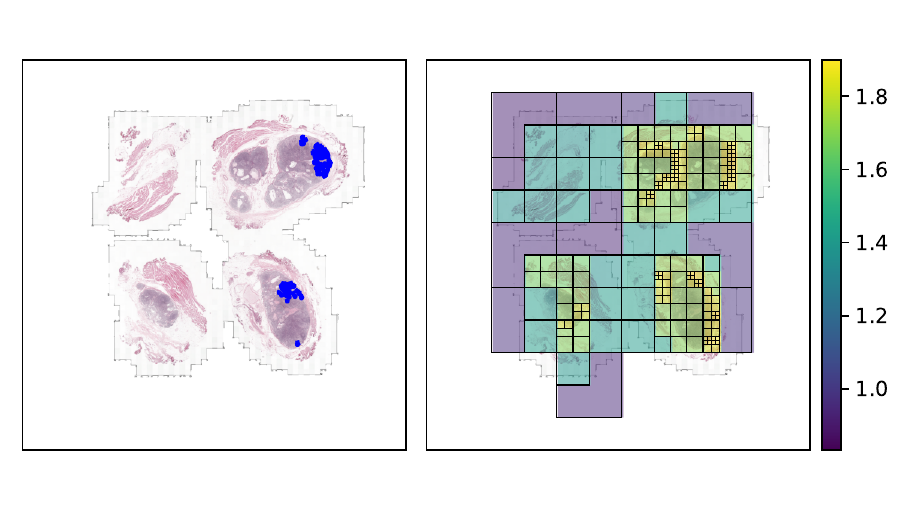}
    \caption{}
\end{subfigure}
\hfill
\begin{subfigure}[t]{0.5\textwidth}
    \includegraphics[trim={0 55 0 55},clip,width=\textwidth]{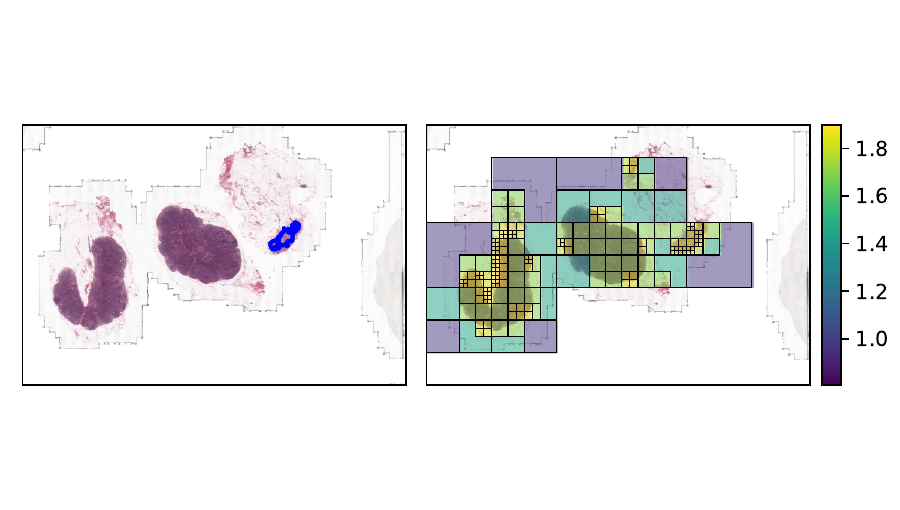}
    \caption{}
\end{subfigure}
\hfill
\begin{subfigure}[t]{0.5\textwidth}
    \includegraphics[trim={0 10 0 10},clip,width=\textwidth]{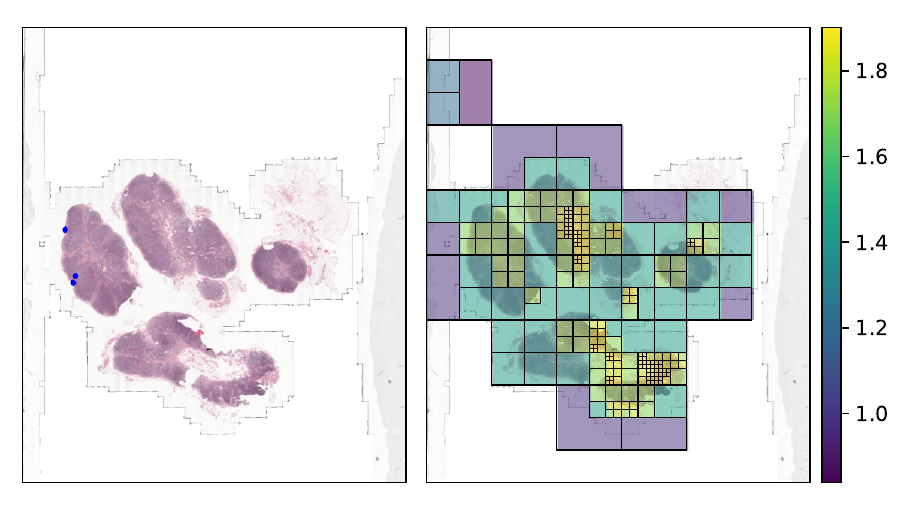}
    \caption{}
\end{subfigure}
\caption{Left: whole slide images from the CAMELYON17 dataset with human-annotated tumours regions marked in blue. Right: visualisation of the patches selected by \name{} across magnifications 0.625x through 10x, and their corresponding importance values. (a) and (b) show strong coverage of the tumorous regions at all magnifications, although (c) shows that \name{} may fail to identify micrometastases in some challenging cases.}
\label{fig:attention_vis}
\end{figure}

\hackyparagraph{Interpretability} The quantification of patch importance (via $\alpha$) enables model interpretability through the explicit identification of regions of interest. \Cref{fig:attention_vis} visualises the patches selected by \name{} on three CAMELYON17 \cite{CAMELYON16_17} slides, alongside manually annotated tumour regions. Note that \name{} \textit{was not trained} on CAMELYON17, but instead applied in a zero-shot setting, following training on TCGA-BRCA. 

The `heat' value for each pixel is given by the sum of the encapsulating patch importances at each magnification level, with $\alpha^i$ weighted by factor of $1/2^i$ to prevent excessive heat in the areas selected across all magnification levels. \name{} appears to correctly identify tumorous regions, and avoids adipose tissue and areas of low tissue density -- despite receiving only weak slide-level supervision. As a small and fixed number of patches ($\npatch=20$) are retained at each magnification level, it is to be expected that not all of the tumorous tissue is selected, and indeed that less relevant patches may be selected in the absence of tumour. \looseness -1

\section{Discussion} \label{sec:discussion}

State-of-the-art methods in computational pathology generally rely on processing entire whole slide images as thousands of patches at high magnification. In this work, we present an alternative approach, in which we filter the processed data to a small subset of relevant patches across several magnification levels. While the recent work of \citet{thandiackal2022zoommil} explores a similar motivation, we approach this problem from the perspective of improving the efficiency of hierarchical processing, rather than extending MIL to multiple magnifications via learnable patch selection,
leading to a more expressive and performant \mbox{(non-MIL)} model that views each patch \textit{in context}. We note that, unlike ZoomMIL, our patch selection algorithm is not differentiable (due to the top-K operation within $\textsc{Filter}$), but we do not find this necessary for strong performance, instead learning $\alpha$ via a gating mechanism. Despite processing strictly less data than most baselines we compare to, \name{} achieves superior performance on average across five large cancer datasets. While we evaluate on survival prediction tasks, \name{} is applicable to arbitrary weakly-supervised tasks and other large-scale image data.

Our ablation study (\Cref{tab:ablation_without_stuff}) demonstrates that incorporating data from multiple magnification levels, here in the form of context, is beneficial for performance. While high magnification patches allow the modelling of cellular-level features of the slide, patches at lower magnification provide convenient representations of higher-level features of the slide, such as the general organisation of the tissue. Our work therefore supports the hypothesis that performance may be improved through the incorporation of patches across magnification levels, as suggested by past work~\cite{multi_res_MIL, Li2020DualstreamMI}. \looseness -1 

\name{} leverages UNI, which like most domain specific patch encoding models, was trained exclusively on patches at high magnification power ($20\times$). However, \name{} requires the encoding of patches across a range of magnifications, including patches at very low magnification, and we therefore hypothesise that performance may be further improved using a \textit{cross-magnification} pre-trained patch encoder. \looseness -1 

Our results support the hypothesis that processing large numbers of patches is often unnecessary for achieving strong performance on practically relevant tasks such as survival prediction.
In fact, by reducing the number of patches input to \name{} we obtained \textit{improved} performance. It allows for the unimpeded use of a transformer architecture (usually restricted by huge sequence lengths), significantly lower memory requirements, faster training times, and an improved signal-to-noise ratio (by excluding patches of low relevance). However, the thin margin between random and attentional patch selection, as observed in \Cref{tab:ablation_without_stuff}, indicates room for improvement in this area, which we leave to future work.

The modelling of explicit important values $\imp$ is an additional benefit of our approach, and highly relevant in a clinical setting.
Despite receiving only weak slide-level supervision, \name{} is capable of identifying important (tumorous) areas of the slide. This capability allows for valuable insights into the model's behaviour, which may ultimately lead to better understanding of the disease. \looseness -1

\section{Conclusion}

We provide strong evidence in this paper to suggest that the processing of entire whole slide images at full magnification is needlessly expensive. Through our design of a novel, patch efficient algorithm, we avoid many of the issues of processing entire slides (high computational cost, poor signal-to-noise ratio, very high latency in practice), improving both efficiency and accuracy. Finally, we demonstrate the benefit that patch contextualisation and slide-level context provide to our unconventional non-MIL approach, and we hope that our work inspires future work in this direction.\looseness-1

    \small
    \bibliographystyle{ieeenat_fullname}
    \bibliography{arxiv}

\begin{thebibliography}{35}
\providecommand{\natexlab}[1]{#1}
\providecommand{\url}[1]{\texttt{#1}}
\expandafter\ifx\csname urlstyle\endcsname\relax
  \providecommand{\doi}[1]{doi: #1}\else
  \providecommand{\doi}{doi: \begingroup \urlstyle{rm}\Url}\fi

\bibitem[Amgad et~al.(2019)Amgad, Elfandy, Hussein, Atteya, Elsebaie, Abo~Elnasr, Sakr, Salem, Ismail, Saad, and et~al.]{BCSS}
Mohamed Amgad, Habiba Elfandy, Hagar Hussein, Lamees~A Atteya, Mai A~T Elsebaie, Lamia~S Abo~Elnasr, Rokia~A Sakr, Hazem S~E Salem, Ahmed~F Ismail, Anas~M Saad, and et al.
\newblock {Structured crowdsourcing enables convolutional segmentation of histology images}.
\newblock \emph{Bioinformatics}, 35\penalty0 (18):\penalty0 3461--3467, 2019.

\bibitem[Carbonneau et~al.(2018)Carbonneau, Cheplygina, Granger, and Gagnon]{MIL_survey_2018}
Marc-André Carbonneau, Veronika Cheplygina, Eric Granger, and Ghyslain Gagnon.
\newblock Multiple instance learning: A survey of problem characteristics and applications.
\newblock \emph{Pattern Recognition}, 77:\penalty0 329--353, 2018.

\bibitem[Caron et~al.(2021)Caron, Touvron, Misra, J\'egou, Mairal, Bojanowski, and Joulin]{DINO}
Mathilde Caron, Hugo Touvron, Ishan Misra, Herv\'e J\'egou, Julien Mairal, Piotr Bojanowski, and Armand Joulin.
\newblock Emerging properties in self-supervised vision transformers.
\newblock In \emph{Proceedings of the International Conference on Computer Vision (ICCV)}, 2021.

\bibitem[Carreira et~al.(2022)Carreira, Koppula, Zoran, Recasens, Ionescu, Henaff, Shelhamer, Arandjelovic, Botvinick, Vinyals, et~al.]{hierarchical_perceiver}
Joao Carreira, Skanda Koppula, Daniel Zoran, Adria Recasens, Catalin Ionescu, Olivier Henaff, Evan Shelhamer, Relja Arandjelovic, Matt Botvinick, Oriol Vinyals, et~al.
\newblock Hierarchical perceiver.
\newblock \emph{arXiv preprint arXiv:2202.10890}, 2022.

\bibitem[Chen et~al.(2021)Chen, Lu, Weng, Chen, Williamson, Manz, Shady, and Mahmood]{multimodal_coattn_transformer}
Richard~J. Chen, Ming~Y. Lu, Wei-Hung Weng, Tiffany~Y. Chen, Drew~F.K. Williamson, Trevor Manz, Maha Shady, and Faisal Mahmood.
\newblock Multimodal co-attention transformer for survival prediction in gigapixel whole slide images.
\newblock In \emph{Proceedings of the IEEE/CVF International Conference on Computer Vision (ICCV)}, pages 4015--4025, 2021.

\bibitem[Chen et~al.(2022)Chen, Chen, Li, Chen, Trister, Krishnan, and Mahmood]{wsi_hierarchical_vit}
Richard~J. Chen, Chengkuan Chen, Yicong Li, Tiffany~Y. Chen, Andrew~D. Trister, Rahul~G. Krishnan, and Faisal Mahmood.
\newblock Scaling vision transformers to gigapixel images via hierarchical self-supervised learning.
\newblock In \emph{Proceedings of the IEEE/CVF Conference on Computer Vision and Pattern Recognition (CVPR)}, pages 16144--16155, 2022.

\bibitem[Chen et~al.(2024)Chen, Ding, Lu, Williamson, Jaume, Chen, Zhang, Shao, Song, Shaban, et~al.]{chen2024uni}
Richard~J Chen, Tong Ding, Ming~Y Lu, Drew~FK Williamson, Guillaume Jaume, Bowen Chen, Andrew Zhang, Daniel Shao, Andrew~H Song, Muhammad Shaban, et~al.
\newblock Towards a general-purpose foundation model for computational pathology.
\newblock \emph{Nature Medicine}, 2024.

\bibitem[Cordonnier et~al.(2021)Cordonnier, Mahendran, Dosovitskiy, Weissenborn, Uszkoreit, and Unterthiner]{cordonnier2021differentiable}
Jean-Baptiste Cordonnier, Aravindh Mahendran, Alexey Dosovitskiy, Dirk Weissenborn, Jakob Uszkoreit, and Thomas Unterthiner.
\newblock Differentiable patch selection for image recognition.
\newblock In \emph{Proceedings of the IEEE/CVF Conference on Computer Vision and Pattern Recognition}, pages 2351--2360, 2021.

\bibitem[Deng et~al.(2009)Deng, Dong, Socher, Li, Li, and Fei-Fei]{ImageNet}
Jia Deng, Wei Dong, Richard Socher, Li-Jia Li, K. Li, and Li Fei-Fei.
\newblock Imagenet: A large-scale hierarchical image database.
\newblock \emph{2009 IEEE Conference on Computer Vision and Pattern Recognition}, pages 248--255, 2009.

\bibitem[Dimitriou et~al.(2019)Dimitriou, Arandjelović, and Caie]{WSIDeepLearningSurvey}
Neofytos Dimitriou, Ognjen Arandjelović, and Peter~D. Caie.
\newblock Deep learning for whole slide image analysis: An overview.
\newblock \emph{Frontiers in Medicine}, 6, 2019.

\bibitem[Dosovitskiy et~al.(2021)Dosovitskiy, Beyer, Kolesnikov, Weissenborn, Zhai, Unterthiner, Dehghani, Minderer, Heigold, Gelly, Uszkoreit, and Houlsby]{ViT}
Alexey Dosovitskiy, Lucas Beyer, Alexander Kolesnikov, Dirk Weissenborn, Xiaohua Zhai, Thomas Unterthiner, Mostafa Dehghani, Matthias Minderer, Georg Heigold, Sylvain Gelly, Jakob Uszkoreit, and Neil Houlsby.
\newblock An image is worth 16x16 words: Transformers for image recognition at scale.
\newblock \emph{ICLR}, 2021.

\bibitem[Han et~al.(2022)Han, Yang, Huang, Zhang, and Yao]{MultimodalDynamics}
Zongbo Han, Fan Yang, Junzhou Huang, Changqing Zhang, and Jianhua Yao.
\newblock Multimodal dynamics: Dynamical fusion for trustworthy multimodal classification.
\newblock In \emph{Proceedings of the IEEE/CVF Conference on Computer Vision and Pattern Recognition (CVPR)}, pages 20707--20717, 2022.

\bibitem[He et~al.(2015)He, Zhang, Ren, and Sun]{ResNet}
Kaiming He, X. Zhang, Shaoqing Ren, and Jian Sun.
\newblock Deep residual learning for image recognition.
\newblock \emph{2016 IEEE Conference on Computer Vision and Pattern Recognition (CVPR)}, pages 770--778, 2015.

\bibitem[He et~al.(2024)He, Wang, Zeng, Liang, Duan, Yang, Pan, He, Huang, and Guan]{MIL_strange_patch_subsampling}
Qiming He, Chengjiang Wang, Siqi Zeng, Zhendong Liang, Hufei Duan, Jingying Yang, Feiyang Pan, Yonghong He, Wenting Huang, and Tian Guan.
\newblock Registration-enhanced multiple instance learning for cervical cancer whole slide image classification.
\newblock \emph{International Journal of Imaging Systems and Technology}, 34\penalty0 (1):\penalty0 e22952, 2024.

\bibitem[Ilse et~al.(2018)Ilse, Tomczak, and Welling]{ABMIL}
Maximilian Ilse, Jakub Tomczak, and Max Welling.
\newblock Attention-based deep multiple instance learning.
\newblock In \emph{Proceedings of the 35th International Conference on Machine Learning}, pages 2127--2136. PMLR, 2018.

\bibitem[Ilse et~al.(2020)Ilse, Tomczak, and Welling]{MIL_pathology}
Maximilian Ilse, Jakub~M. Tomczak, and Max Welling.
\newblock Deep multiple instance learning for digital histopathology.
\newblock In \emph{Handbook of Medical Image Computing and Computer Assisted Intervention}, pages 521--546. Academic Press, 2020.

\bibitem[Li et~al.(2020{\natexlab{a}})Li, Li, and Eliceiri]{Li2020DualstreamMI}
Bin Li, Yin Li, and Kevin~W. Eliceiri.
\newblock Dual-stream multiple instance learning network for whole slide image classification with self-supervised contrastive learning.
\newblock \emph{2021 IEEE/CVF Conference on Computer Vision and Pattern Recognition (CVPR)}, pages 14313--14323, 2020{\natexlab{a}}.

\bibitem[Li et~al.(2020{\natexlab{b}})Li, Li, Sisk, Ye, Wallace, Speier, and Arnold]{multi_res_MIL}
Jiayun Li, Wenyuan Li, Anthony~E. Sisk, Huihui Ye, William~D. Wallace, W. Speier, and Corey~W. Arnold.
\newblock A multi-resolution model for histopathology image classification and localization with multiple instance learning.
\newblock \emph{Computers in biology and medicine}, 131:\penalty0 104253, 2020{\natexlab{b}}.

\bibitem[Li et~al.(2018)Li, Yao, Zhu, Li, and Huang]{Li2018GraphCF}
Ruoyu Li, Jiawen Yao, Xinliang Zhu, Yeqing Li, and Junzhou Huang.
\newblock Graph cnn for survival analysis on whole slide pathological images.
\newblock In \emph{International Conference on Medical Image Computing and Computer-Assisted Intervention}, 2018.

\bibitem[Litjens et~al.(2018)Litjens, Bandi, Ehteshami Bejnordi, Geessink, Balkenhol, Bult, Halilovic, Hermsen, van de Loo, Vogels, Manson, Stathonikos, Baidoshvili, van Diest, Wauters, van Dijk, and van der Laak]{CAMELYON16_17}
Geert Litjens, Peter Bandi, Babak Ehteshami Bejnordi, Oscar Geessink, Maschenka Balkenhol, Peter Bult, Altuna Halilovic, Meyke Hermsen, Rob van de Loo, Rob Vogels, Quirine~F Manson, Nikolas Stathonikos, Alexi Baidoshvili, Paul van Diest, Carla Wauters, Marcory van Dijk, and Jeroen van der Laak.
\newblock 1399 h\&amp;e-stained sentinel lymph node sections of breast cancer patients: the camelyon dataset.
\newblock \emph{GigaScience}, 7\penalty0 (6):\penalty0 giy065, 2018.

\bibitem[Liu et~al.(2021)Liu, Lin, Cao, Hu, Wei, Zhang, Lin, and Guo]{Swin}
Ze Liu, Yutong Lin, Yue Cao, Han Hu, Yixuan Wei, Zheng Zhang, Stephen Lin, and Baining Guo.
\newblock Swin transformer: Hierarchical vision transformer using shifted windows.
\newblock In \emph{Proceedings of the IEEE/CVF International Conference on Computer Vision (ICCV)}, 2021.

\bibitem[Lu et~al.(2020)Lu, Williamson, Chen, Chen, Barbieri, and Mahmood]{MIL_ImageNet}
Ming~Y. Lu, Drew F.~K. Williamson, Tiffany~Y. Chen, Richard~J. Chen, Matteo Barbieri, and Faisal Mahmood.
\newblock Data-efficient and weakly supervised computational pathology on whole-slide images.
\newblock \emph{Nature Biomedical Engineering}, 5:\penalty0 555 -- 570, 2020.

\bibitem[Otsu(1979)]{OtsusMethod}
Nobuyuki Otsu.
\newblock A threshold selection method from gray-level histograms.
\newblock \emph{IEEE Transactions on Systems, Man, and Cybernetics}, 9\penalty0 (1):\penalty0 62--66, 1979.

\bibitem[Pocock et~al.(2022)Pocock, Graham, Vu, Jahanifar, Deshpande, Hadjigeorghiou, Shephard, Bashir, Bilal, Lu, Epstein, Minhas, Rajpoot, and Raza]{tiatoolbox}
Johnathan Pocock, Simon Graham, Quoc~Dang Vu, Mostafa Jahanifar, Srijay Deshpande, Giorgos Hadjigeorghiou, Adam Shephard, Raja Muhammad~Saad Bashir, Mohsin Bilal, Wenqi Lu, David Epstein, Fayyaz Minhas, Nasir~M Rajpoot, and Shan E~Ahmed Raza.
\newblock {TIAToolbox as an end-to-end library for advanced tissue image analytics}.
\newblock \emph{Communications Medicine}, 2\penalty0 (1):\penalty0 120, 2022.

\bibitem[Ronneberger et~al.(2015)Ronneberger, Fischer, and Brox]{UNet}
Olaf Ronneberger, Philipp Fischer, and Thomas Brox.
\newblock U-net: Convolutional networks for biomedical image segmentation.
\newblock In \emph{Medical Image Computing and Computer-Assisted Intervention -- MICCAI 2015}, pages 234--241, Cham, 2015. Springer International Publishing.

\bibitem[Shao et~al.(2021)Shao, Bian, Chen, Wang, Zhang, Ji, et~al.]{TransMIL}
Zhuchen Shao, Hao Bian, Yang Chen, Yifeng Wang, Jian Zhang, Xiangyang Ji, et~al.
\newblock Transmil: Transformer based correlated multiple instance learning for whole slide image classification.
\newblock \emph{Advances in Neural Information Processing Systems}, 34:\penalty0 2136--2147, 2021.

\bibitem[Simonyan and Zisserman(2014)]{VGG16}
Karen Simonyan and Andrew Zisserman.
\newblock Very deep convolutional networks for large-scale image recognition.
\newblock \emph{CoRR}, abs/1409.1556, 2014.

\bibitem[Thandiackal et~al.(2022)Thandiackal, Chen, Pati, Jaume, Williamson, Gabrani, and Goksel]{thandiackal2022zoommil}
Kevin Thandiackal, Boqi Chen, Pushpak Pati, Guillaume Jaume, Drew~FK Williamson, Maria Gabrani, and Orcun Goksel.
\newblock Differentiable zooming for multiple instance learning on whole-slide images.
\newblock In \emph{The European Conference on Computer Vision (ECCV)}, 2022.

\bibitem[Vaswani et~al.(2017)Vaswani, Shazeer, Parmar, Uszkoreit, Jones, Gomez, Kaiser, and Polosukhin]{Transformer}
Ashish Vaswani, Noam~M. Shazeer, Niki Parmar, Jakob Uszkoreit, Llion Jones, Aidan~N. Gomez, Lukasz Kaiser, and Illia Polosukhin.
\newblock Attention is all you need.
\newblock In \emph{Neural Information Processing Systems}, 2017.

\bibitem[Wulczyn et~al.(2020)Wulczyn, Steiner, Xu, Sadhwani, Wang, Flament-Auvigne, Mermel, Chen, Liu, and Stumpe]{MIL_with_random_patch_selection}
Ellery Wulczyn, David~F. Steiner, Zhaoyang Xu, Apaar Sadhwani, Hongwu Wang, Isabelle Flament-Auvigne, Craig~H. Mermel, Po-Hsuan~Cameron Chen, Yun Liu, and Martin~C. Stumpe.
\newblock Deep learning-based survival prediction for multiple cancer types using histopathology images.
\newblock \emph{PLOS ONE}, 15\penalty0 (6):\penalty0 1--18, 2020.

\bibitem[Xiong et~al.(2021)Xiong, Zeng, Chakraborty, Tan, Fung, Li, and Singh]{nystrom}
Yunyang Xiong, Zhanpeng Zeng, Rudrasis Chakraborty, Mingxing Tan, Glenn~Moo Fung, Yin Li, and Vikas Singh.
\newblock Nystr{\"o}mformer: A nystr{\"o}m-based algorithm for approximating self-attention.
\newblock \emph{Proceedings of the ... AAAI Conference on Artificial Intelligence. AAAI Conference on Artificial Intelligence}, 35 16:\penalty0 14138--14148, 2021.

\bibitem[Yao et~al.(2020)Yao, Zhu, Jonnagaddala, Hawkins, and Huang]{DeepAttnMIL}
Jiawen Yao, Xinliang Zhu, Jitendra Jonnagaddala, Nicholas~J Hawkins, and Junzhou Huang.
\newblock Whole slide images based cancer survival prediction using attention guided deep multiple instance learning networks.
\newblock \emph{Medical image analysis}, 65:\penalty0 101789, 2020.

\bibitem[Zadeh and Schmid(2021)]{NLL_Loss}
Shekoufeh~Gorgi Zadeh and Matthias Schmid.
\newblock Bias in {Cross-Entropy-Based} training of deep survival networks.
\newblock \emph{IEEE Trans Pattern Anal Mach Intell}, 43\penalty0 (9):\penalty0 3126--3137, 2021.

\bibitem[Zhang et~al.(2022)Zhang, Zhang, Zhao, Chen, Arik, and Pfister]{nested_hierarchical_transformer}
Zizhao Zhang, Han Zhang, Long Zhao, Ting Chen, Sercan~O. Arik, and Tomas Pfister.
\newblock Nested hierarchical transformer: Towards accurate, data-efficient and interpretable visual understanding.
\newblock In \emph{AAAI Conference on Artificial Intelligence (AAAI)}, 2022.

\bibitem[Zhao et~al.(2020)Zhao, Yang, Fang, Liu, Zhou, Zhang, Sun, Yang, Menze, Fan, and Yao]{Zhao2020PredictingLN}
Yu Zhao, Fan Yang, Yuqi Fang, Hailing Liu, Niyun Zhou, Jun Zhang, Jiarui Sun, Sen Yang, Bjoern~H Menze, Xinjuan Fan, and Jianhua Yao.
\newblock Predicting lymph node metastasis using histopathological images based on multiple instance learning with deep graph convolution.
\newblock \emph{2020 IEEE/CVF Conference on Computer Vision and Pattern Recognition (CVPR)}, pages 4836--4845, 2020.

\end{thebibliography}

\clearpage
\setcounter{page}{1}
\maketitlesupplementary
\appendix






\section{\name{} Algorithm}
\label{app:algorithms}
\Cref{alg:processor} describes the algorithm carried out by each processor, and \Cref{alg:slide_proc} the overall \name{} algorithm applied to a slide $X$, in which $\textsc{Predict}$ denotes the final linear layer whose output is the prediction~$\hat{y}$.

\begin{algorithm}[ht]
\caption{Processing algorithm for patches at magnification $m_i$}\label{alg:processor}
\begin{algorithmic}[1]
\Procedure{$\proc_i$}{$\Xms{m_i}, \C(\Xms{m_i})$}
    \State $\Yms{m_i} \longleftarrow \imenc(\Xms{m_i}) + \text{RNN}(\C(\Xms{m_i}))$
    \State $\alpha^i \longleftarrow \textsc{Imp}_i(\Yms{m_i})$
    \State $\Zms{m_i} \longleftarrow \alpha^i \odot \Yms{m_i}$
    \State $F^i \longleftarrow \textsc{GlobalAgg}_i(\Zms{m_i})$
    \State \textbf{return} $(F^i, \alpha^i)$
\EndProcedure
\end{algorithmic}
\end{algorithm}

\begin{algorithm}[ht]
\caption{Overall \name{} slide processing algorithm}\label{alg:slide_proc}
\begin{algorithmic}[1]
\Procedure{Process}{$X$}
    \State $\Xms{m_1} \longleftarrow \Xm{m_1}$
    \State $\Cs \longleftarrow [~]$
   \For {$i=1$ to $\nmag$}
        \State $(F^i, \alpha^i) \longleftarrow \proc_i(\Xms{m_i}, \C(\Xms{m_i}))$
        \State $\Xms{m_{i+1}} \longleftarrow \textsc{Magnify}(\textsc{Filter}(\Xms{m_i}, \alpha^i))$
        \State $\Cs \longleftarrow \Cs + [F^i]$
    \EndFor
\State \textbf{return} \textsc{Predict}($\Cs$)
\EndProcedure
\end{algorithmic}
\end{algorithm}
We may formally define $\textsc{Filter}$ as
\[
\textsc{Filter}(\Xms{m_i}, \alpha^i) = \left\{\Xms{m_i}_{u,v} \mid (u,v)\in\textsc{TopK}(\alpha^i)\right\},
\]
where $\textsc{TopK}$ returns the patch coordinates $(u_1,v_1)$, $(u_2,v_2), \dots$, $(u_\npatch, v_\npatch)$ of the top $\npatch$ values in $\alpha^i$. Note that, although $\Xms{m_i}$ is an indexed set, we use set notation for readability.
We may formally define $\textsc{Magnify}$ as
\[
\begin{split}
\textsc{Magnify}(\tilde{U}^{m_i}) = \biggl\{\Xms{m_{i+1}}_{u,v} \mid & \left(\floor{\frac{u}{\magf}},\floor{\frac{v}{\magf}}\right) \in dom(\tilde{U}^{m_i}) \\
&  \land \textsc{HasTissue}(\Xms{m_{i+1}}_{u,v}) \biggr\},
\end{split}
\]
where $\tilde{U}^{m_i}$ is the (indexed set) output of $\textsc{Filter}$, $dom$ extracts the index from its input (here, the patch coordinates contained in $\tilde{U}^{m_i}$), and $\textsc{HasTissue}$ returns whether the input patch contains above a certain threshold of tissue, as identified using Otsu's method \cite{OtsusMethod}.


\section{Hyperparameters} \label{app:hyperparameters}
\Cref{tab:hyperparameters} gives the hyperparameters chosen for \name{}, and \Cref{tab:zoommil_hyperparameters} the hyperparameters chosen for ZoomMIL, which we choose to enable fair comparison between our methods. We initially used 40 epochs for ZoomMIL to match \name{}, but found that performance was improved when training for 100 as in the sample configuration.

\begin{table}[ht]
    \centering
    \caption{\name{} hyperparameters, shared between all datasets.}
    \begin{tabular}{lr}
        \toprule
        Hyperparameter & Value     \\
        \midrule
        Learning rate & 2e-5     \\
        Batch size & 32\\
        Epochs & 40     \\
        Survival quantisation bins ($b$) & 4     \\
        Censored data loss weight $\alpha$ & 0.6\\
        \midrule
        Image encoder & UNI \cite{chen2024uni} \\
        Patch size & 256 \\
        Patches extracted per level $\npatch$ & 20 \\
        Magnification factor $M$ & 2 \\
        Hierarchy depth $\nmag$ & 5 (from $0.625\times$ to $10\times$) \\
        Transformer aggregator dimension & 128\\
        Transformer aggregator heads & 4\\
        Transformer aggregator layers & 2\\
        $\textsc{IMP}_i$ hidden dimension & 128\\
        LSTM hidden dimension & 256\\
        \bottomrule
    \end{tabular}
    
    \label{tab:hyperparameters}
\end{table}

\begin{table}[ht]
    \centering
      \caption{ZoomMIL hyperparameters, shared between all datasets. We use their public implementation, adding support for survival prediction. The configuration was taken from the sample configuration provided in the repository, with $K$ changed to match our configuration. \textdagger{} the ZoomMIL codebase only supports a batch size of 1 and hierarchy depth of 3. The magnifications $1.25\times$, $2.5\times$, $10\times$ were chosen to match the configuration used for BRIGHT in the original paper \cite{thandiackal2022zoommil}.}
    \begin{tabular}{lr}
        \toprule
        Hyperparameter & Value     \\
        \midrule
        Learning rate & 1e-4     \\
        Batch size & 1\textsuperscript{\textdagger}\\
        Epochs & 100     \\
        Survival quantisation bins ($b$) & 4     \\
        Censored data loss weight $\alpha$ & 0.6\\
        \midrule
        Image encoder & UNI \cite{chen2024uni} \\
        Patch size & 256 \\
        Patches extracted per level $\npatch$ & 20 \\
        Hierarchy depth $\nmag$ & 3\textsuperscript{\textdagger} ($1.25\times$, $2.5\times$, $10\times$) \\
        $\sigma$ & 0.002 \\
        \bottomrule
    \end{tabular}
  
    \label{tab:zoommil_hyperparameters}
\end{table}

\section{Inference Speed Experiment Details} \label{app:inference_speed_details}
This section details the method used to produce \Cref{fig:inference_speed}.

\textbf{ABMIL} First, background patches are identified using Otsu's method \cite{OtsusMethod}, applied to a low resolution version of the WSI. The patches are then loaded sequentially and processed using UNI in batches of 256 on the GPU.

\textbf{\name{}} Patches are loaded sequentially, and processed with UNI in a single batch per magnification level (as there are a small number at each level). Otsu's method is used in the $\textsc{Filter}$ function to identify background patches. 

\begin{figure}
    \centering
    \includegraphics[width=\linewidth]{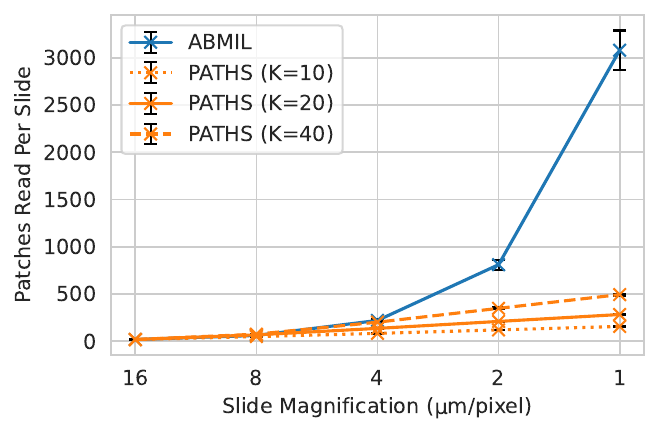}
    \caption{Number of patches loaded per slide for ABMIL (blue) compared to \name{} (orange) for various values of $\npatch$. Values averaged over 50 slides from TCGA-BRCA, as with \Cref{fig:inference_speed}. Unlike inference latency, this measure is not hardware dependent, and demonstrates clearly the exponential growth in the number of patches required by traditional MIL approaches compared to the linear number required by \name{}.}
    \label{fig:npatches_mil_vs_ours}
\end{figure}

In both cases, patches are loaded from disk using a single CPU thread, and pre-processing and model inference takes place on a single A100 GPU. In all experiments, loading and pre-processing the patches using UNI takes over 99\% of the total time (and over 99.99\% for the example MIL model), despite \name{} loading fewer patches. \Cref{fig:npatches_mil_vs_ours} gives the corresponding number of patches loaded by each method, providing a measure of inference efficiency independent of hardware or image encoder.

\section{Choice of Image Encoder}\label{app:image_encoder_ablation}

\Cref{tab:im_enc_table} compares the performance of \name{} for three different image encoders. To evaluate the importance of domain specific encoding for \name{}, we compare using UNI to using an ImageNet pre-trained ResNet50, and observe weaker performance in the latter case. We then compare two models trained specifically on WSI patches, a self-supervised vision transformer trained on a number of TCGA datasets at $20\times$ magnification \cite{wsi_hierarchical_vit} and UNI, a vision transformer trained on patches across many different tissue types, also at $20\times$ \cite{chen2024uni}. As discussed in \Cref{sec:discussion}, \name{} processes patches at magnifications strictly less than $20\times$ in our experiments -- between $0.625\times$ and $10\times$ -- which are out of domain inputs for both models. Due to its larger scale, high-resolution fine-tuning, and exposure to a larger number of tissue types, UNI appears to create superior representations of these low magnification patches, leading to stronger performance. We therefore select UNI as our image encoder. However, we emphasise that performance of \name{} may be improved further through the use of an image encoder pre-trained on WSI patches at lower magnification.

\begin{table*}[ht]
    \centering
    \advance\leftskip-3cm
    \advance\rightskip-3cm
       \caption{C-index performance of \name{} for three different choices of image encoder $\imenc$: ResNet50 pre-trained on ImageNet (RN50) \cite{ResNet, ImageNet}, a vision transformer trained in a self-supervised manner on WSI patches at 20x (SSL-ViT) \cite{wsi_hierarchical_vit, DINO}, and UNI \cite{chen2024uni}. The results highlight the insufficiency of ImageNet pre-trained patch encoders on pathology tasks, and the poor quality of the low magnification patch features produced by SSL-ViT.}
    \begin{tabular}{l|ccccc|c}
         \toprule
         Image Encoder $\imenc$ & IDC & CRC & CCRCC & PRCC & LUAD & Mean \\
         \midrule
         RN50 \cite{ResNet, ImageNet} & $0.590\pm0.093$ & $0.612\pm0.056$ & $0.596\pm0.053$ & $0.619\pm0.125$ & $0.490\pm0.058$ & 0.581 \\
         SSL-ViT \cite{wsi_hierarchical_vit} & $0.575 \pm 0.066$ & $0.516 \pm 0.016$ & $0.583 \pm 0.054$ & $0.545 \pm 0.103$ & $\bf 0.547 \pm 0.044$ & 0.553 \\
         UNI \cite{chen2024uni} & $\bf 0.636\pm0.069$ & $\bf 0.695\pm0.097$ & $\bf 0.677\pm0.046$ & $\bf 0.772\pm0.036$ & $0.545\pm0.060$ & \bf 0.665 \\
         \bottomrule
    \end{tabular}
 
    \label{tab:im_enc_table}
\end{table*}

\section{Further Visualisations} 
\begin{figure*}
    \centering
    \includegraphics[width=\linewidth]{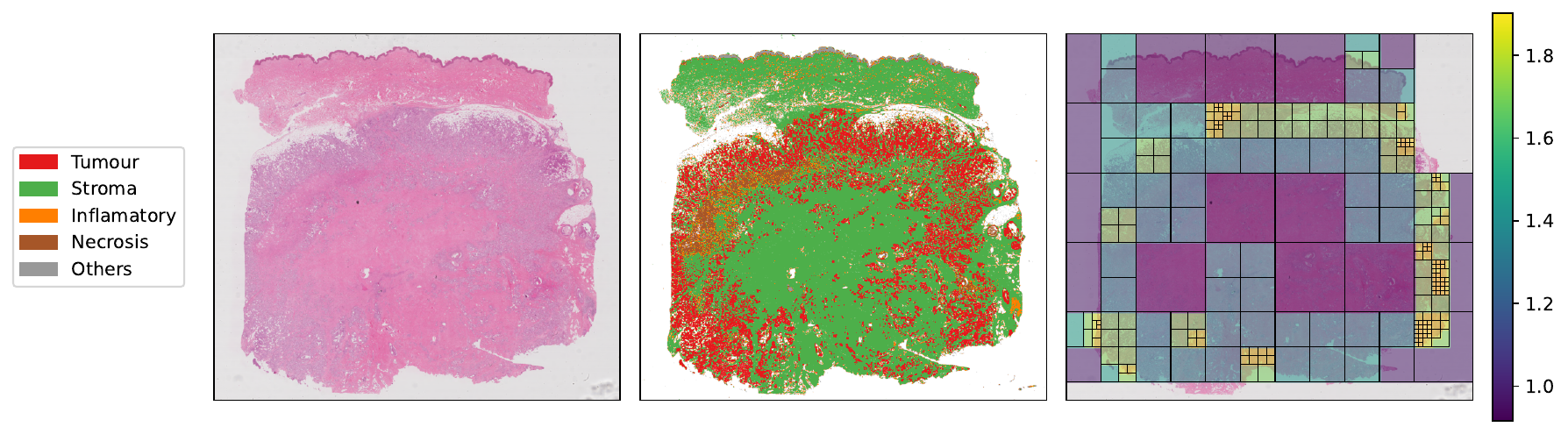}
    \caption{Left-to-right: whole slide image from TCGA-BRCA, predicted semantic segmentation, \name{} heatmap. 
    We observe that \name{} appears to focus on tumorous regions.}
    \label{fig:brca_heatmap}
\end{figure*}

\Cref{fig:brca_heatmap} shows an example heatmap of \name{} on TCGA-BRCA. Due to a lack of ground truth labels in TCGA, we display the predicted semantic segmentation alongside the \name{} heatmap. The prediction was computed using a U-Net model provided by \verb|tiatoolbox|, pre-trained on the BCSS dataset (an annotated patch-level dataset derived from TCGA-BRCA) \cite{UNet, BCSS, tiatoolbox}. Due to the computational cost of evaluating this model, which requires the extraction and processing of tens of thousands of patches at $20\times$ magnification per slide, and lack of human annotated labels, we provide one such example only.

\begin{figure*}[ht]
    \centering
    \begin{tabular}{cc}
        \begin{subfigure}{0.48\textwidth}
            \centering
            \includegraphics[trim={0 25 0 25},clip,width=\textwidth]{figs/HM3_patient_010_node_4.pdf}
            \caption{}
        \end{subfigure} &
        \begin{subfigure}{0.48\textwidth}
            \centering
            \includegraphics[trim={0 55 0 55},clip,width=\textwidth]{figs/HM3_patient_015_node_2.pdf}
            \caption{}
        \end{subfigure} \\
        \begin{subfigure}{0.48\textwidth}
            \centering
            \includegraphics[trim={0 10 0 10},clip,width=\textwidth]{figs/HM3_patient_004_node_4.pdf}
            \caption{}
        \end{subfigure} &
        \begin{subfigure}{0.48\textwidth}
            \centering
            \includegraphics[trim={0 5 0 5},clip,width=\textwidth]{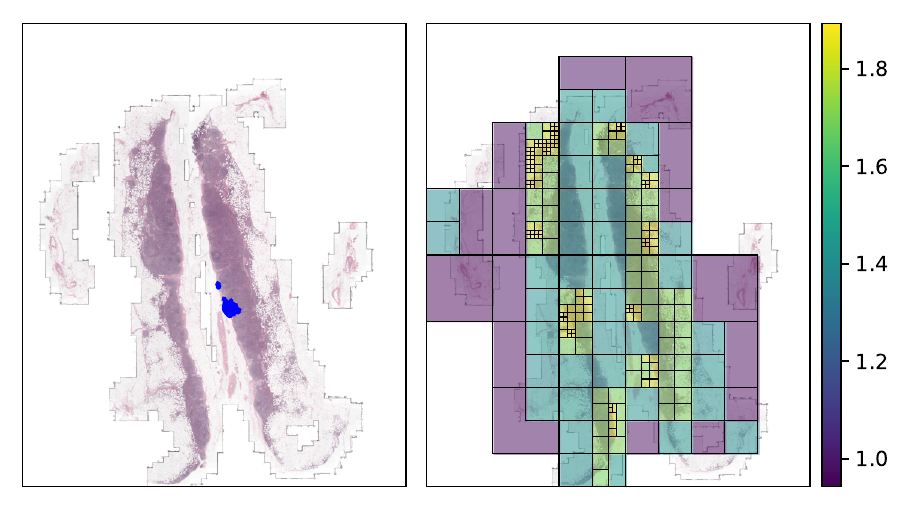}
            \caption{}
        \end{subfigure} \\
        \begin{subfigure}{0.48\textwidth}
            \centering
            \includegraphics[trim={0 25 0 25},clip,width=\textwidth]{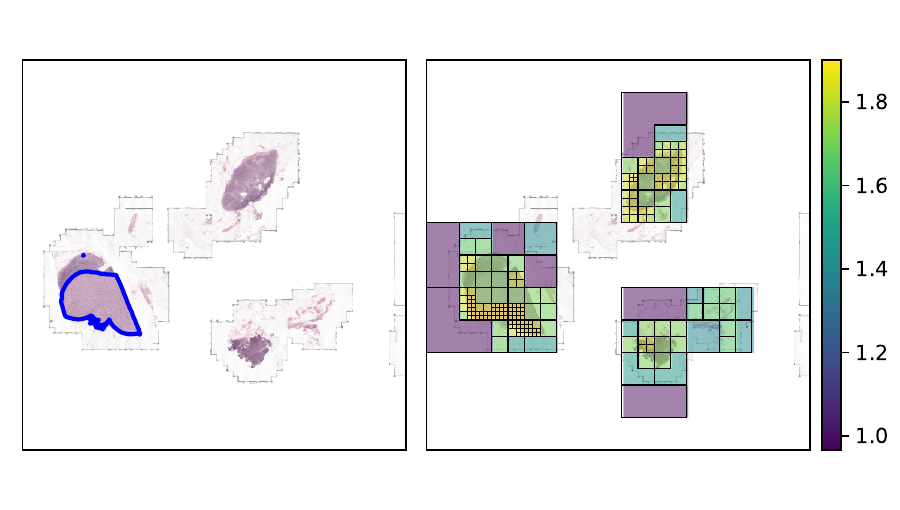}
            \caption{}
        \end{subfigure} &
        \begin{subfigure}{0.48\textwidth}
            \centering
            \includegraphics[trim={0 40 0 40},clip,width=\textwidth]{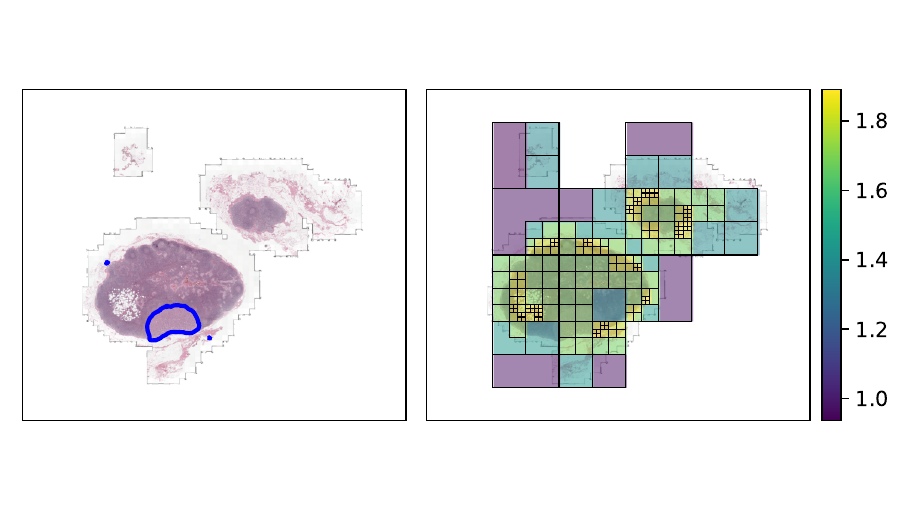}
            \caption{}
        \end{subfigure} \\
        \begin{subfigure}{0.48\textwidth}
            \centering
            \includegraphics[trim={0 25 0 25},clip,width=\textwidth]{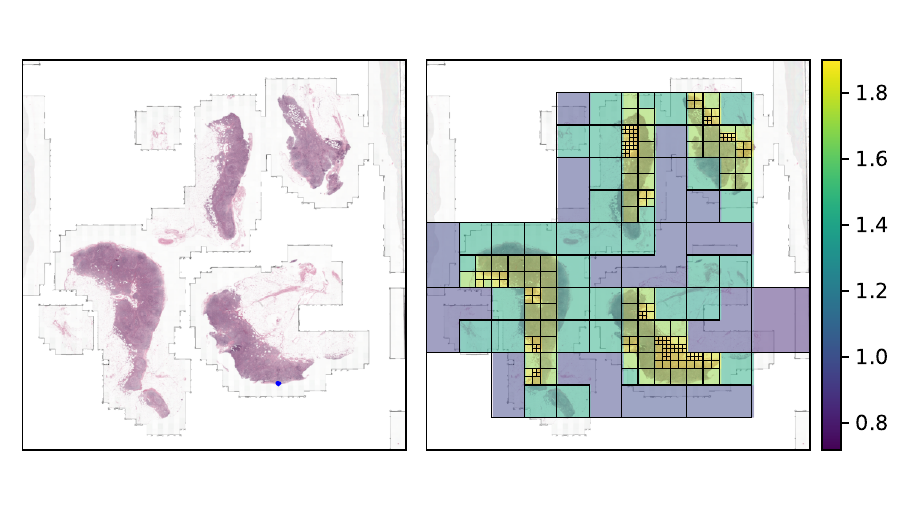}
            \caption{}
        \end{subfigure} &
        \begin{subfigure}{0.48\textwidth}
            \centering
            \includegraphics[trim={0 25 0 25},clip,width=\textwidth]{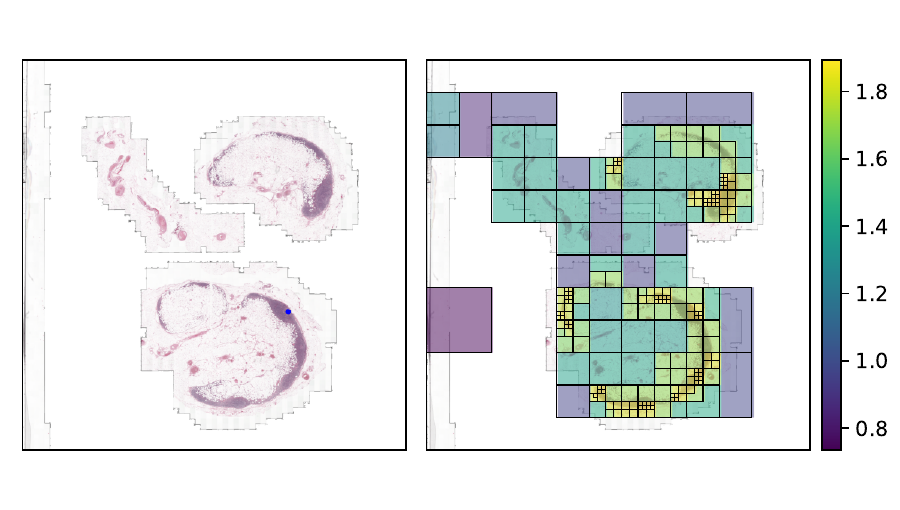}
            \caption{}
        \end{subfigure} \\
    \end{tabular}
\caption{Further region of interest examples from CAMELYON17, with human-annotated tumourous regions marked in blue on the left of each figure, and the patches selected by our zero-shot \name{} model on the right. (a), (b) and (c) correspond to \Cref{fig:attention_vis}. To prevent cherry picking, this figure contains the first eight slides alphabetically of the 50 annotated slides in CAMELYON17. Though \name{} appears to miss micrometastases in (c) and (g), the other examples show strong coverage of tumorous regions despite the zero-shot application of \name{}. We also highlight that \name{} visibly avoids selecting adipose tissue, particularly in (b) and (h).}
\label{fig:attention_vis_supp}
\end{figure*}

\Cref{fig:attention_vis_supp} displays further examples from CAMELYON17 \cite{CAMELYON16_17}. Both figures use the \name{} model trained on TCGA-BRCA with seed 0, applied in-domain in \Cref{fig:brca_heatmap}, and zero-shot to CAMELYON17 in \Cref{fig:attention_vis_supp}.

\end{document}